\documentclass[letterpaper]{article}
\usepackage{aaai}
\usepackage{times}
\usepackage{helvet}
\usepackage{courier}
\usepackage{graphicx}
\usepackage{color}
\usepackage{float}
\usepackage{placeins}
\usepackage{subfig}
\usepackage{amsmath}
\usepackage{amsfonts}
\usepackage[utf8]{inputenc}
\usepackage[english]{babel}
\usepackage{natbib}
\usepackage{multirow}
\nocopyright
\title{Recreating Bat Behavior on Quad-rotor UAVs---A Simulation Approach}
\author{M. Hassan Tanveer$^\ast$, Antony Thomas$^\dagger$, Xiaowei Wu$^\ast$, Rolf M\"uller$^\ddag$, Pratap Tokekar$^\S$,  Hongxiao Zhu$^\ast$\\
$^\ast$Department of Statistics, Virginia Tech, USA\\
$^\dagger$DIBRIS, University of Genoa, Italy\\
$^\ddag$Department of Mechanical Engineering, Virginia Tech, USA\\
$^\S$Department of Computer Science, University of Maryland, USA}
\begin{document}

\setlength\parindent{0pt}

\maketitle
\begin{abstract}
\begin{quote}
We develop an effective computer model to simulate sensing environments that consist of natural trees. The simulated environments are random and contain full geometry of the tree foliage. While this simulated model can be used as a general platform for studying the sensing mechanism of different flying species, our ultimate goal is to build bat-inspired Quad-rotor UAVs---UAVs that can recreate bat's flying behavior (e.g., obstacle avoidance, path planning) in dense vegetation. To this end, we also introduce an foliage echo simulator that can produce simulated  echoes by mimicking bat's biosonar.  In our current model, a few realistic model choices or assumptions are made. First, in order to create natural looking trees, the branching structures of trees are modeled by L-systems, whereas the detailed geometry of branches, sub-branches and leaves is created by randomizing a reference tree in a CAD object file. Additionally, the foliage echo simulator is simplified so that no shading effect is considered. We demonstrate our developed model by simulating real-world scenarios with multiple trees and compute the corresponding impulse responses along a Quad-rotor trajectory. 
\end{quote}
\end{abstract}

\section{Introduction}

Many environments, such as dense vegetation and narrow caves, are not easily accessible by human beings. Unmanned Aerial Vehicles (UAVs) provide cost-effective alternatives to human beings for a large variety of tasks in such environments, including search, rescue, surveillance, and land inspection. In recent years, impressive progress has been made in UAVs, leading to revolutions in the aerodynamic structure, mechanical transmission, actuator, computer control, etc. Despite these advances, existing technology in UAVs is still limited as most systems can only operate in clear, open space \cite{Dey2011} or in fields with sparsely distributed tree obstacles \cite{Barry2017}, and most existing approaches for localization and planning fail in the presence of large number of obstacles. Moreover, sensors used in these systems are often bulky which hinders efficient navigation \cite{abdallah2019reliability}. It is highly desirable to build safe and efficient UAV systems that do not fail under different real-world conditions.

Among many directions in technological innovation, bio-inspired technology provides a promising solution that may break the performance boundary in UAVs.  Mammals, insects and other organisms often exhibit advanced capabilities and features that would be desirable for UAVs. They can rapidly pick out salient features buried in large amounts of data, and adapt themselves to the dynamics of their environments. Adopting prototypes that emulate the characteristics and functions found in living creatures may enable robots to maneuver more efficiently without the aid of approaches such as simultaneous mapping and localization (SLAM), GPS or inertial units. In recent years, bio-inspired approaches have already given rise to robots that operate in water \cite{yao2011applications}, air \cite{duan2014pigeon} and on land \cite{zhou2012survey} and, in some cases, transiting in various media. 
 %Among various types of UAVs, Quad-rotor UAVs have received increasing attention due to their vertical takeoff/landing capability and their hovering ability compared with fixed wing UAVs \cite{kim2009accurate}.
 For UAVs in particular, ``Microbot" has been created in 2002 by The California Institute of Technology \cite{bogue2015miniature}, which achieves independent fly by imitating the morphological properties of versatile bat wings. In 2011, AeroVironment successfully developed the ``Hummingbird" by mimicking hummingbirds  \cite{coleman2015design}. The Hummingbird is trained and equipped to continue flying itself with its own supply of energy. The flapping wings can effectively control its attitude angles. Besides these examples, there are several other conventional designs developed, including Robird \cite{folkertsma2017robird}, DelFly \cite{de2016delfly}, and Bat Bot \cite{ramezani2015bat}.

In this research, we consider using the echolocation system of bats as a biological model for the study of highly parsimonious biosonar sensors for UAVs. Millions of years' biological development provides bats numerous incredible skills to navigate freely in complex, unstructured environments. Relying on miniature sonar systems with a few transducers---a nose (or mouth) and two ears, bats achieve much better navigation performance than engineered systems. Specifically, a echolocating bat emits brief ultrasonic pulses through mouth or nostrils, and use the returning echoes to navigate \cite{Griffin1958}.  Based on bats' biosonar, we aim to develop a bat-inspired sonar sensing and navigation paradigm for quad-rotor UAVs. To achieve this, we adopt a data-driven approach that integrates large-scale simulations with statistical learning to gain insights and replicate bats’ abilities.

\begin{figure}[h]
\hspace{-0.5cm}
  \includegraphics[width=1.2\linewidth]{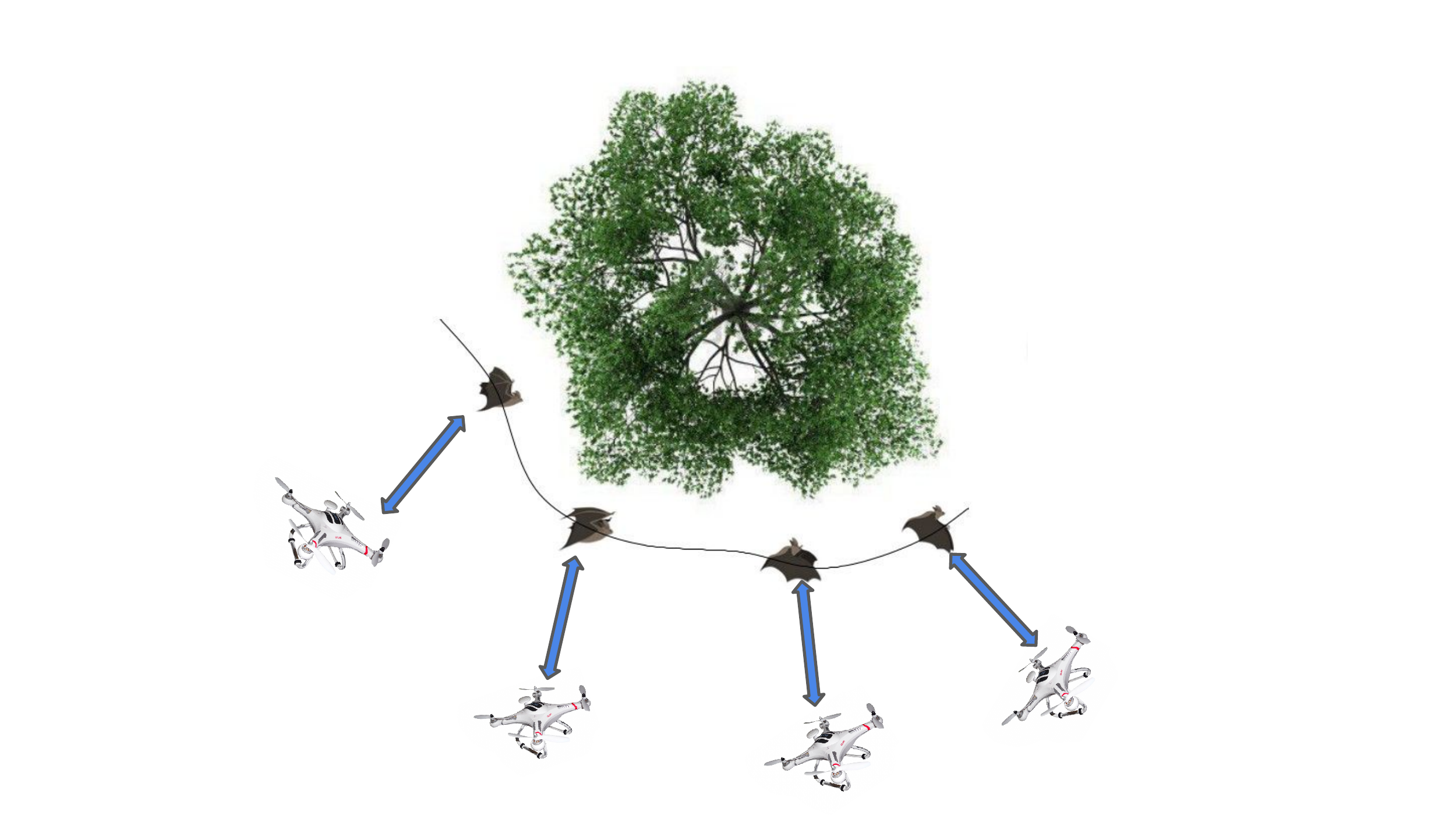}
  \caption{A bat navigating around a tree. We mimic the highly developed bio-sonar system in bats by simulating sonar and leaf beampatterns and validate it through different experiments. A Quad-rotor using this sonar is visualized.}
  \label{fig1}
\end{figure}

Results presented in this paper are based on our initial efforts in recreating the sensory world of bats via computer simulation. We develop an effective computer model to simulate sensing environments that consist of natural looking trees. The simulated environments are random and contain full geometry of the tree foliage. While this model can be used as a general platform for studying the sensing mechanism of different flying species, our ultimate goal is to build bat-inspired Quad-rotor UAVs---UAVs that can recreate bat's flying behavior (e.g., obstacle avoidance, path planning) in dense vegetation. To this end, we also introduce an foliage echo simulator that can produce simulated echoes by mimicking bat's biosonar. In Figure \ref{fig1}, we demonstrate how a bat is mimicked by a Quad-rotor while navigating across a tree. %we create a simulated forest environment mimicking real trees and study the foliage echoes returned by the bio-sonar to extract relevant environment details.
In our current model, a few realistic model choices or assumptions are made. First, in order to create natural looking trees, the branching structures of trees are modeled by L-systems, whereas the detailed geometry of branches, sub-branches and leaves is created by randomizing a reference tree in a CAD object file. Additionally, the foliage echo simulator is simplified so that no shading effect is considered. We demonstrate our developed model by simulating real-world scenarios with multiple trees and compute the corresponding impulse responses along a Quad-rotor trajectory.

%\begin{figure}
%\hspace{-0.5cm}
  %\includegraphics[width=1.2\linewidth]{figures/batwithtree.pdf}
 % \caption{Principle behind bat sonar emission and echo recognition.}
 % \label{fig2}
%\end{figure}
%To this end, we turn to nature and leverage the mechanism used by bats to perceive, plan and act in dense vegetation. We mimic the highly developed bio-sonar system in bats to detect obstacles and regions of interest, as shown in Figure \ref{fig2}. We create a simulated forest environment mimicking real trees and study the foliage echoes returned by the bio-sonar to extract relevant environment details.

The rest of this paper is organized as follows. In Section 2, we describe the method of simulating a sensing environment with multiple natural looking trees and the theory behind the foliage echo simulator. We elaborate experimental results and analyses in Section 3. Finally, in Section 4, a general conclusion and the direction towards future work are given.

\section{Material and Methods}

%In the recent past, there has been a proliferation of different approaches for autonomous navigation. Yet, most approaches are limited to operating in open space environments with less obstacles. Furthermore, most approaches used image based sensors and hence fares poorly in low-lit conditions. To mitigate these limitations, we propose the use of bio-inspired sonar sensors as an extension of the work in \cite{10.1371/journal.pone.0182824}. As such, our approach comprises of mainly three different stages-- computer simulation, model development, and experimental validation. In the remainder of this Section, we elucidate the different stages and discuss the necessary background for the same. \\
We develop a computational framework that consists of two simulators, one for the simulation of sensing environment which produces random trees with necessary geometry (e.g., leave locations, size and orientations etc.), another for the simulation of  foliage echoes which produces sonar impulses by mimicking the biosonar system of bats. In this Section, we elucidate the main methodology used in these simulators.
\\

\begin{figure}[h]
\hspace{-0.5cm}
  \includegraphics[width=1.2\linewidth]{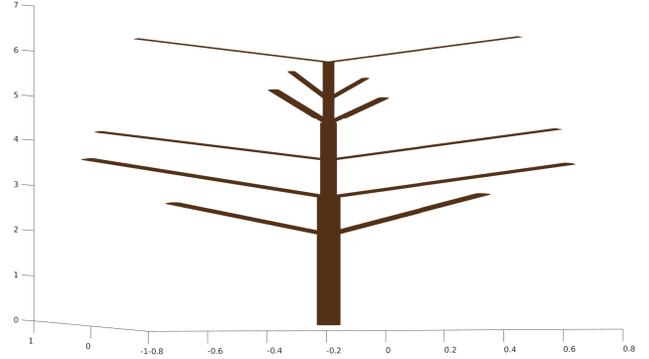}
  \caption{L-system for branch generation on the trunk.}
  \label{fig3}
\end{figure}

We simulate the topology of each individual tree by combining Lindenmayer systems (L-systems) with modified CAD implemented object files. An L-system is a graphical model commonly used to describe the growth pattern of plants \cite{Prusinkiewicz1996}.  It defines the branching pattern of a plant through recursively applying certain production rules on a string of symbols. Each symbol in the string defines a structural component (e.g., branch, terminal). Each recursive iteration creates an additional level of growth of the string. The final string represents the branching structure of the grown tree. While L-system is commonly used to produce branching structures \cite{shlyakhter2001reconstructing}, we found that it is not sufficient for generating natural looking trees because of over-simplified assumptions. For example, tree models based on L-systems often model branches as straight lines while ignoring the natural curvatures of the branches. Furthermore, most of the L-system models often rely on a few parameters to control the lengths, thickness and angles of branches. Although probability distributions can be introduced to randomize these parameters, they are often not enough to characterize all features of a particular tree species. For these reasons, we choose to adopt L-system to generate the first level branching locations at the trunk. To generate the branches and sub-branches, we modify reference trees from CAD developed object files by randomizing the branch curvatures, lengths, and sub-branch locations. This produces random trees that look more realistic. In Figure \ref{fig3}, we demonstrate the plot of a tree simulated by L-system with the first level branching structure only.

\begin{figure}[h]
    \includegraphics[scale=0.3]{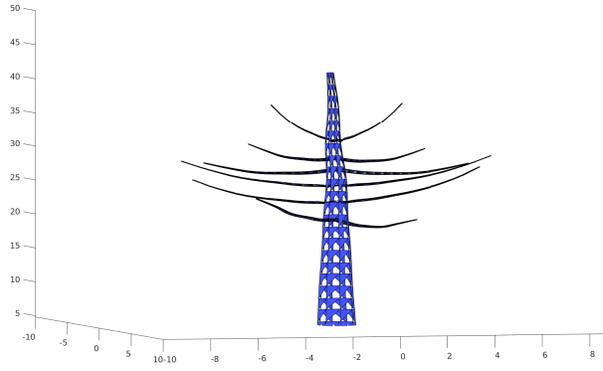}%
    \qquad
    \hspace{-1cm}
    \includegraphics[scale=0.3]{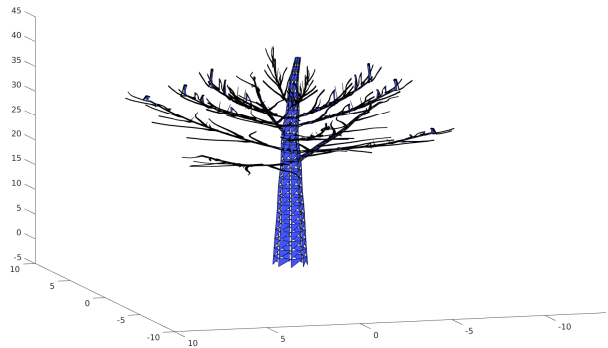}%
    \caption{L-system and object file fusion for generating tree branches and sub branches}%
    \label{fig4}%
\end{figure}

Information about the branches, sub-branches and leaves of the simulated trees is stored as an organizational structure of building systems. This is associated with 3D CAD drawing that includes faces and vertices modelled as meshes. %\cmt{Each branch, sub-branch and leaf contains the axis and attitude of the trunk with a starting point of branch taken from L-system. (rewrite this sentence.)} 
This provides a complete 3D tree with planarity for each branch. The planarity makes it easy to visualize the tree with short computing time based on available data (i.e: Polygon, vertices, textures etc), thereby offering a convenient way to effectively imagine scenarios with other trees in forests.  
%By assuming continuous deformation \cmt{(what this mean?)}, 
%branches can automatically be given 3D structural designs, as shown in Figure \ref{fig4} 
The L-system does not really follow drawing standards (i.e: with the geometric information). Hence, in order to make branches and sub-branches, we should follow certain rules using 3D CAD tools that abides by the tree geometry (see Figure \ref{fig4}).

Based on the simulator of a random tree, we are able to generate a community that consists of random number of trees. We determine the number of trees and the locations of these trees in a 2-D region by sampling from Inhomogeneous Poisson process (IPP). %Since we discuss a simulation based study, it is imperative to simulate real-world conditions that facilitate autonomous navigation in any given environment. We simulate natural environments to generate trees at different locations and their respective branches and leaves. Tree locations in the environment are determined by sampling from an Inhomogeneous Poisson process (IPP) model. 
Let $D \subset \mathbb{R}^2$ denote the 2-D region on which the community of trees will be built. The random locations (i.e., $(x,y)$ coordinates) of the trees are denoted by $S = \{s_i\}_{1\leq i\leq n}$. We assume that $S$ follows an IPP with intensity function $\lambda (s) : D \rightarrow \mathbb{R}^+ $, where $\lambda (s)$ is a parameter to be specified by user which describes how dense the trees are at every location. Small values of $\lambda (s)$ indicate sparse regions whereas high values indicate dense regions. The number of trees, $n$, follows a Poisson distribution $\int_D \lambda (s) ds$. To simulate $S$ given $n$, we adopt a thinning approach \cite{Lewis1979}.\\

For the simulation of foliage echoes, we follow the approach of \cite{10.1371/journal.pone.0182824}. Here, we briefly summarize the method. In the current model, the leaves are simplified as circular disks. 
The simulated foliage echoes are stored as time-domain (discrete) signals. Let $Y = \{y_1,\ldots, y_n\}$ denote one time-domain signal to be simulated. Let $Y^* = \{y_1^*,\ldots, y_{n^{'}}^* \}$ denote the Fourier transform of $Y$ in the frequency domain. To obtain $Y$, we first compute $Y^*$ and apply inverse fast Fourier transform. It is assumed that $y_k^*$ is nonzero in the frequency ranges between 60 to 80 kHz, which corresponds to the strongest harmonic in the biosonar impulses of the \textit{Rhinolophus ferrumequinum bat}~\cite{andrews2003AC}. According to 
acoustic laws of sound reflection \cite{Bowman1987},
each Fourier component $y_k^*$ is the superposition of all the reflecting echoes from the reflecting facets within the main lobe of the sonar. It takes the form %The superposition of all the reflecting echoes from the reflecting facets, for example leaves, has the form

\begin{equation}
    y^*_k = \sum_{i=1}^m A_{ki}\cos(\phi_{ki}) + j \sum_{i=1}^m A_{ki}\sin(\phi_{ki}),
\end{equation}

where $m$ denotes the number of reflecting facets within the main lobe of the sonar, $A_{ki}$ is the amplitude at frequency $f_k$ (which is the frequency corresponding to $y^*_k$) for the $i$-th facet, $\phi_{ki}$ is a phase delay parameter at $f_k$ for the $i$-th facet. The term $A_{ki}$ can be computed by
\begin{equation}
    A_{ki} = S(az_i,el_i,f_k,r_i)L_i(\beta_i,a_i,f_k)\frac{\lambda_k}{2\pi r_i^2},
\end{equation}

where $S(az_i,el_i,f_k,r_i)$ represents the sonar beampattern with $az_i$ and $el_i$ being the azimuth and elevation angles of the line that connects the sonar with the $i$-th reflecting facet, $r_i$ is the distance between the sonar and the $i$-th reflecting facet, $L_i(\beta_i,a_i,f_k)$ is the beampattern of the reflecting facet with $\beta_i$ and $a_i$ being the incident angle and  of the $i$-th reflecting facet respectively. The sonar beampattern has the general form

\begin{multline}
    S(\cdot) = A_{1}\exp\{- ( a(x-x_0)^2 +  \\ 2b(x-x_0)(y-y_0) + c(y-y_0)^2 t)\}
    \label{eq:sbeam}
\end{multline}

where $A_{1}$ is the amplitude, $a, b, c$ are the parameters of Gaussian function. The value of $a, b, c$ are determined by empirical data. The leaf beampattern can be approximated by cosine function of the form

\begin{equation}
    L_i(\cdot) = A\left(c\left(f_k,a_i \right) \cdot\cos \left(B c\left( f_k,a_i\right) \cdot\beta_i\right) \right)
    \label{eq:lbeam}
\end{equation}

where $c = 2 \pi a_i {f_k}/{v}$, with $v$ being the speed of sound and $A$, $B$ are functions of $c$. A detailed description of (\ref{eq:sbeam}) and (\ref{eq:lbeam}) is beyond the scope of this paper and we refer the interested readers to~\cite{bowman1987book,adelman2014arXiv}.

The simulation of sensing environments and foliage echoes provide us rich amount of sensing data under various sensing tasks. Within the simulated setup, we can design a Quad-rotor that mimics the bat behaviors for tasks such as insects prey. 
%After the construction of trees and Bio inspired sonar, the Bat mimicked Quad-rotor carries the Bio inspired sonar sensor, in order to get lift and maneuver around the trees in search for home in forest. 
The dynamics of Quad-rotor UAV can be taken from, e.g., \cite{bouabdallah2007design}. We employ the control, that has been recently developed to stabilize UAV while flying \cite{tanveer_recchiuto_sgorbissa_2019}.

%The statistical model proposed facilitates our interpretation of parsimonious echolocation detection in densely heavily forested habitats. Established sensing systems concentrate on point-by-point image analysis of the situation utilizing cameras or ultrasonic sensors, generating vast amounts of sensor data that seem to be costly to maintain to process with minimal on-board computation.  Our method relies on one-dimensional bio-sonar signals, adopting new statistical training to evaluate the specifications needed for the problem at hand, eliminating the complexity of reconstruction of the entire environment.  The use of a bat-inspired sonar system in computing is a special feature to our strategy.  This bio inspired strategy provides an extra diverse aspect, allowing for parsimonious yet effective sensing.  The research findings can be extended in unstructured or dark environments to multiple platforms for detecting and collision avoidance. 

\section{Results and Discussion}

We performed a pilot study by designing a simple sensing scene that involves multiple trees. These trees are constructed by combing an L-system with CAD developed object files as described in the previous section. When visualizing the trees, leaves were approximated using the mid points of the triangular meshes used to model leaves in CAD.

We conduct several simulations in the MATLAB environment to demonstrate the performance of model. The performance are evaluated on an Intel{\small\textregistered} 
Core$™$ i7-3632QM under Ubuntu 16.04 LTS. The simulation with multiple trees has been done on a ten-core server computer. Tree locations in the environment are determined by sampling from an IPP model. The trees are different from each other in terms of of branches angles, sizes and leaves distribution. Moreover, the initial branching pattern follows that of an L-system. In each simulation, we construct a tree (or trees) and analyze the impulse responses from simulated sonar echoes. The impulse responses are computed at different sonar locations in the environment to mimic a flying Quad-rotor. For example, we have computed impulses at regular intervals along a circular path around a tree and impulses for a trajectory directly towards the tree. The beam width of the sonar main lobe is chosen to be 10, 20, and 50 degrees.% With bandwidth 10, the accuracy is much better 
%\cmt{with good impulse response (what does this mean?)} as compare to wider bandwidths. \cmt{Thus, choosing a smaller bandwidth often provides ...value in order to achieve the maximum existence of different number branches and leaves density (what does this mean).}

%\includegraphics[scale=0.5]{figures/1.eps}
%\includegraphics[scale=0.5]{figures/differentrotors_experiemnt.pdf}

\begin{figure}[h]
\hspace{-0.5cm}
  \includegraphics[width=1.2\linewidth]{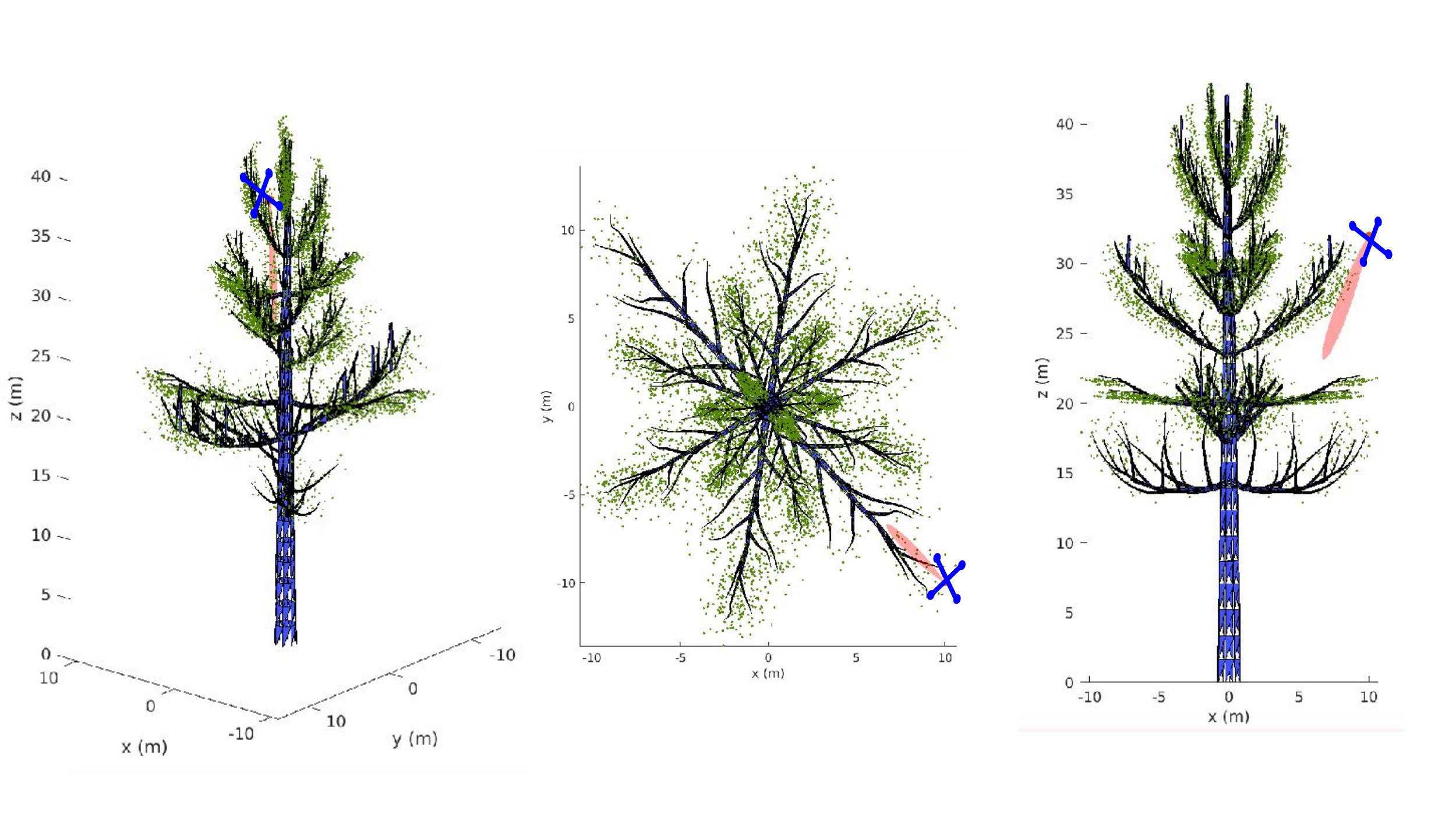}
  \caption{Quad-rotor navigating with no leaves encountering in main lobe of sensor}
  \label{fig5}
\end{figure}

\begin{figure}[h]
\hspace{-0.5cm}
  \includegraphics[width=1.2\linewidth]{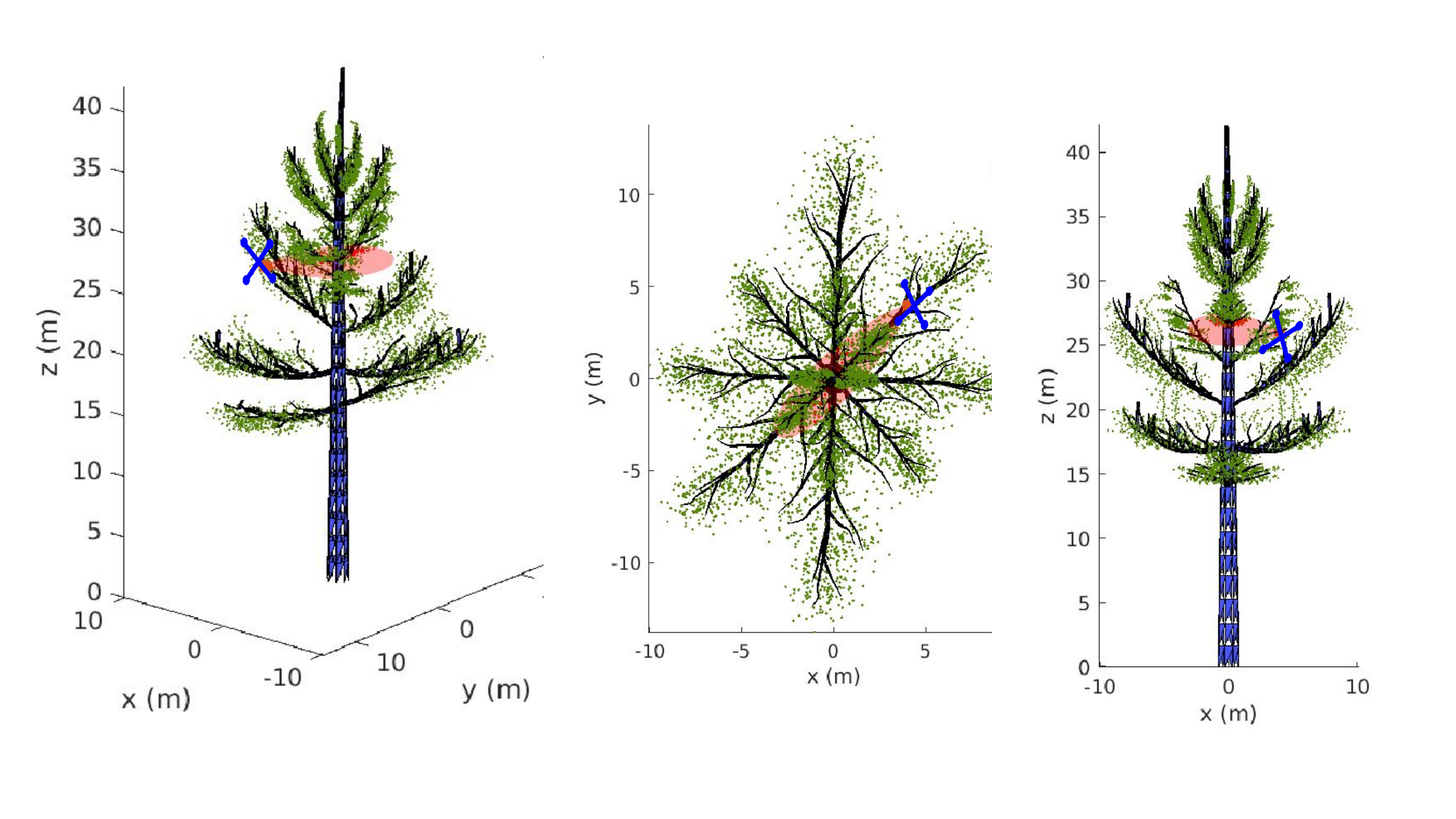}
  \caption{Quad-rotor navigating with leaves encountering in main lobe of sensor}
  \label{fig6}
\end{figure}

\begin{figure}[h]
\hspace{-0.5cm}
  \includegraphics[width=1\linewidth]{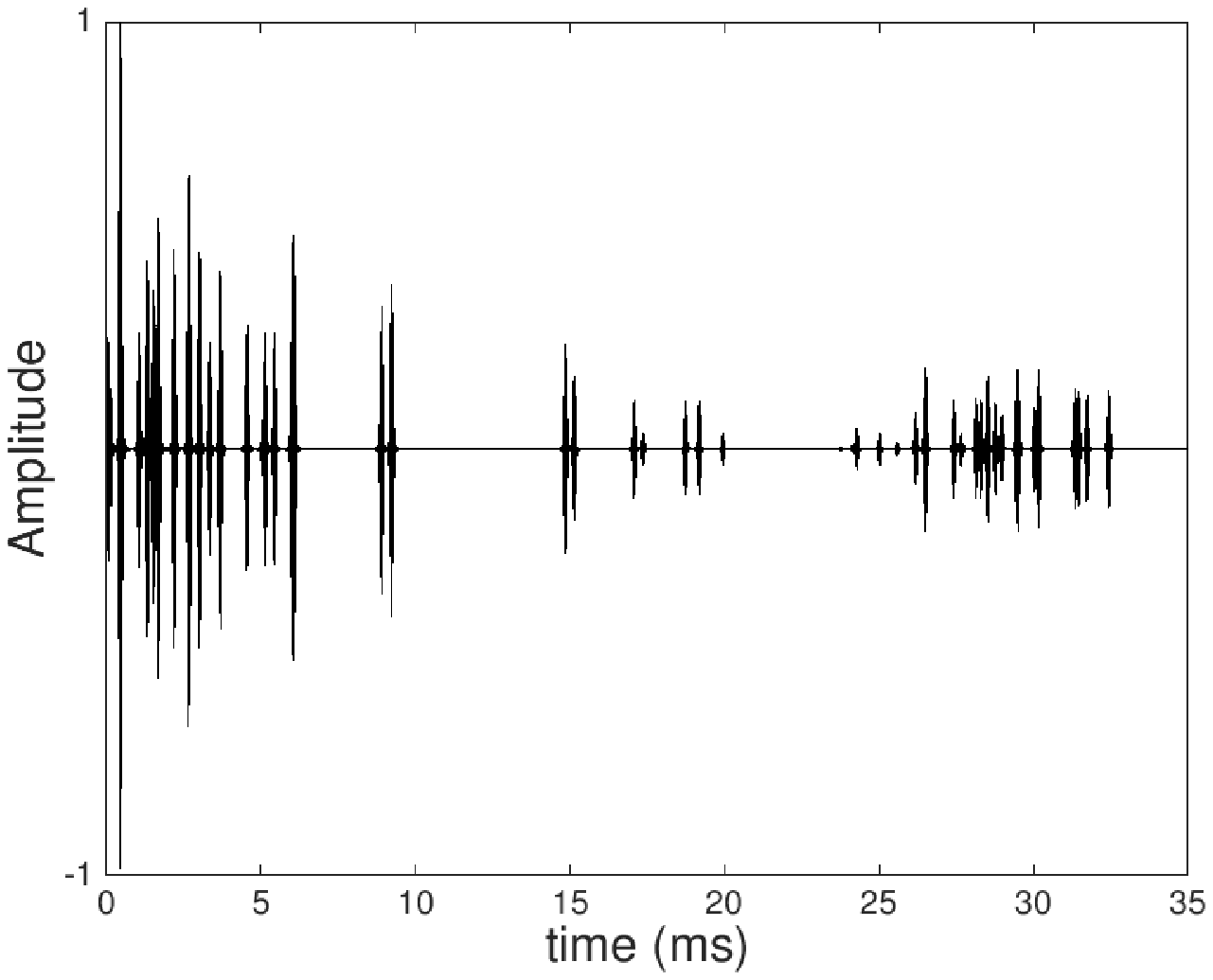}
  \caption{Impulse response of main lobe}
  \label{fig7}
\end{figure}

\begin{figure}[h]
\hspace{-0.5cm}
  \includegraphics[width=1.2\linewidth]{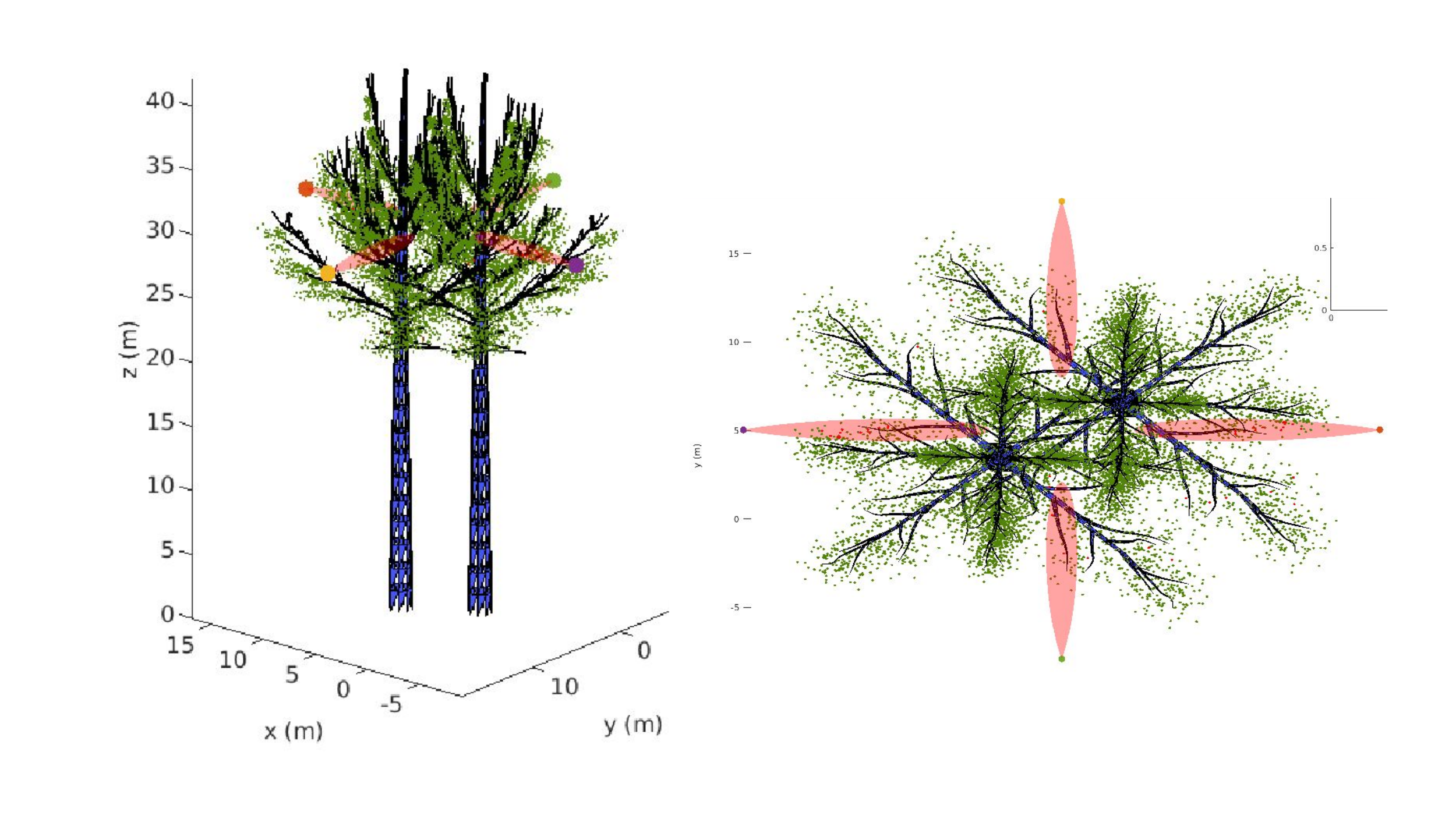}
  \caption{Quad-rotor navigating across 2 trees with leaves encountering in main lobe of sensor}
  \label{fig8}
\end{figure}

\begin{figure}[h]
    \centering
        \subfloat{\includegraphics[scale=0.29]{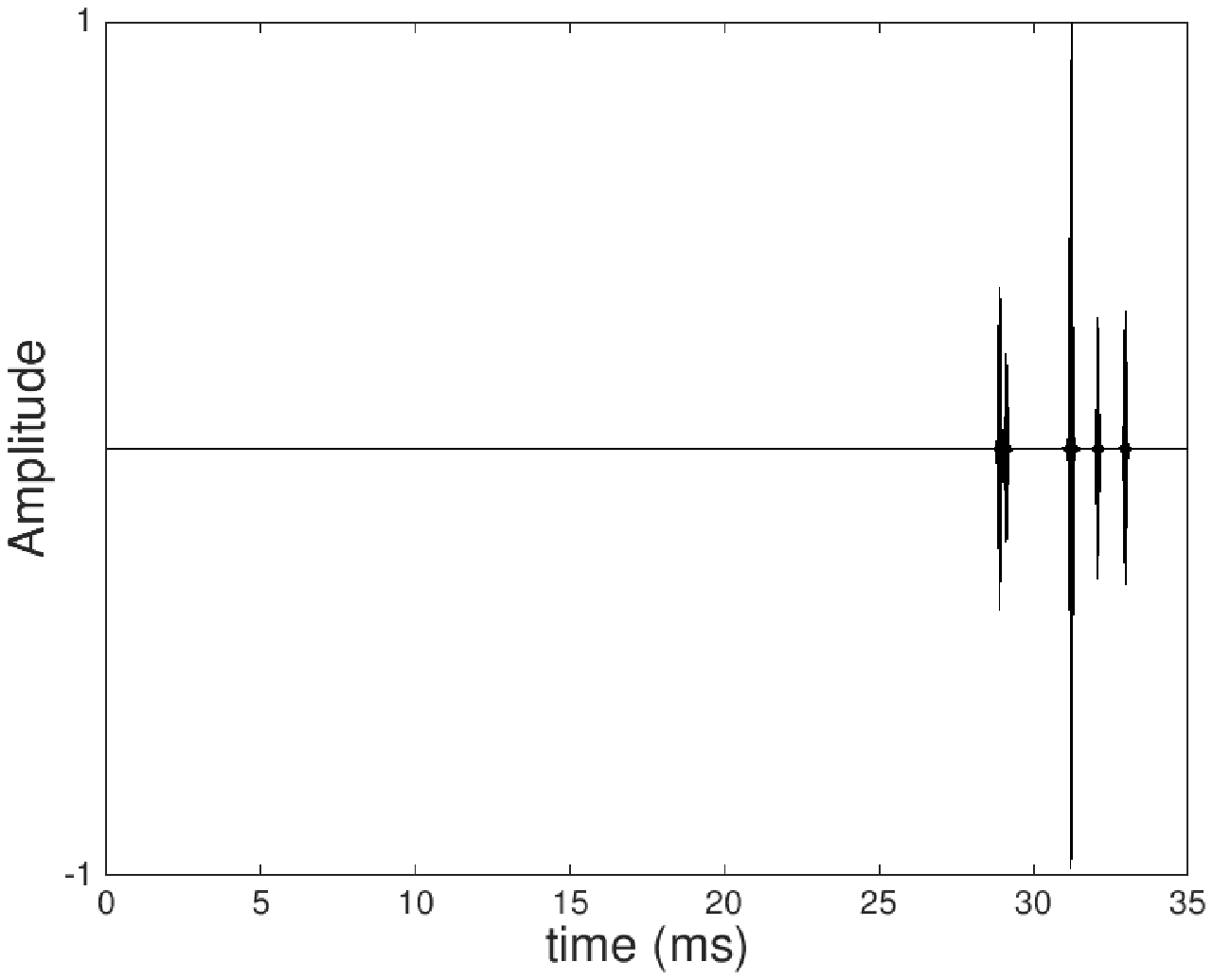}}
\subfloat{\includegraphics[scale=0.29]{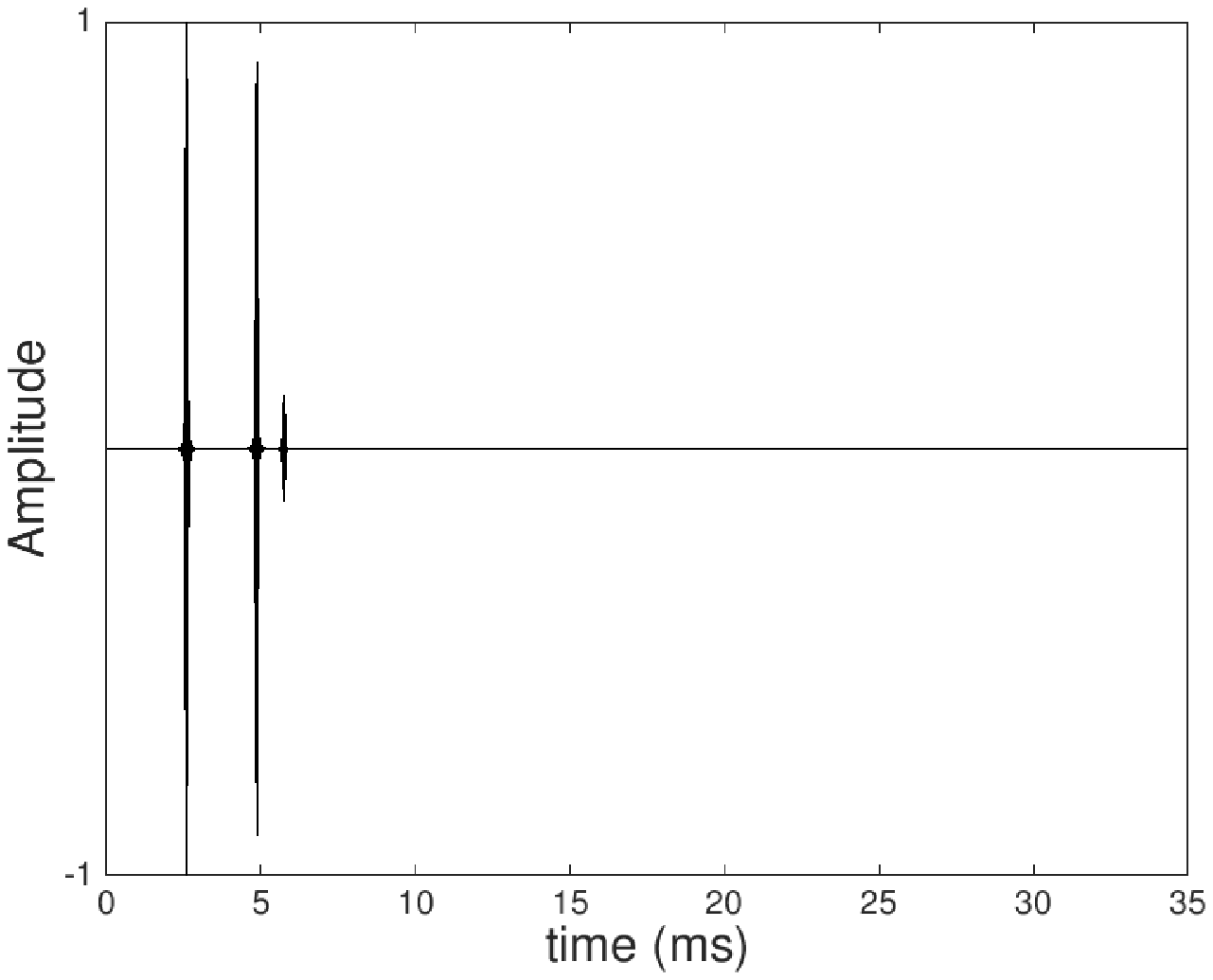}}\\
\subfloat{\includegraphics[scale=0.29]{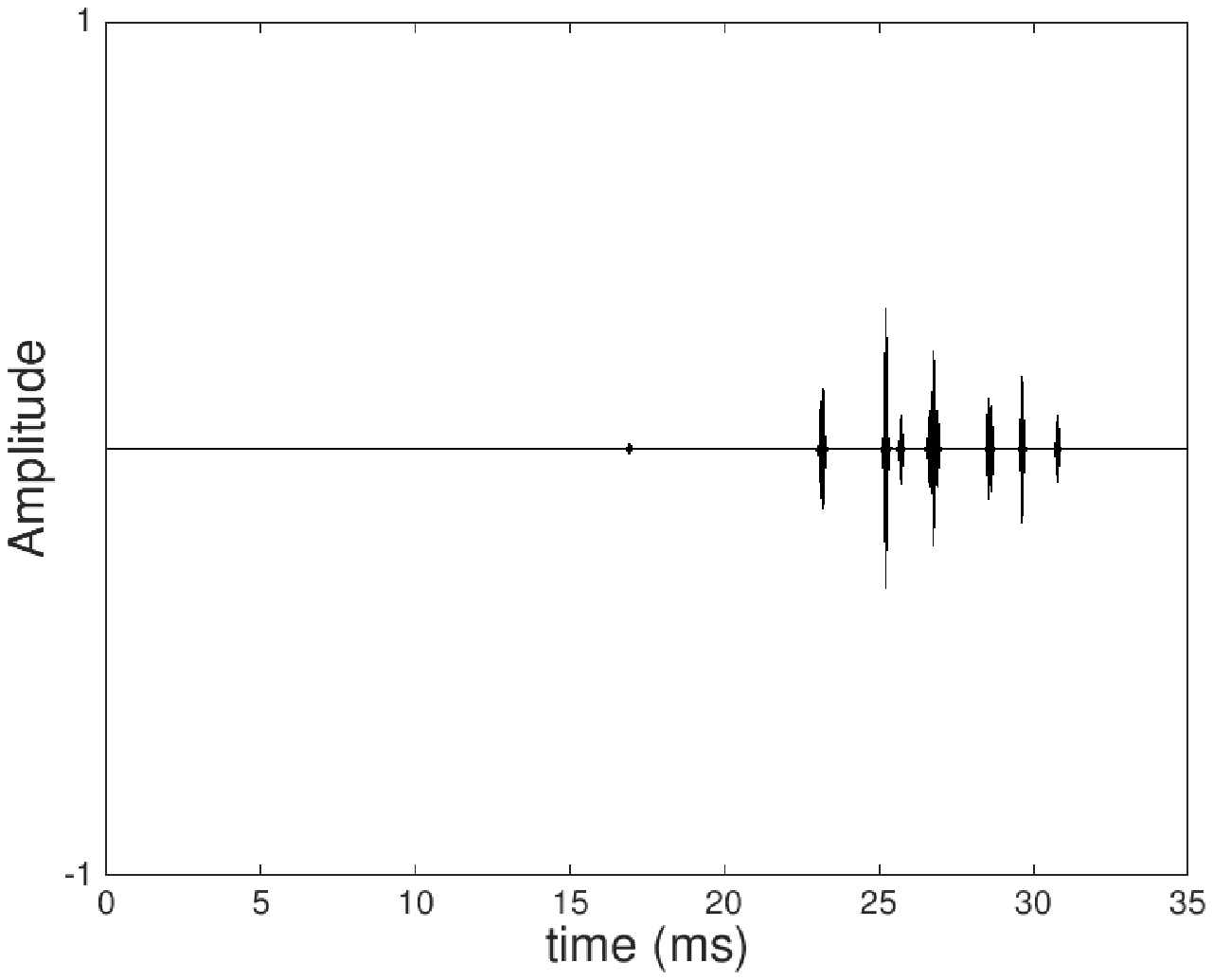}}
\subfloat{\includegraphics[scale=0.29]{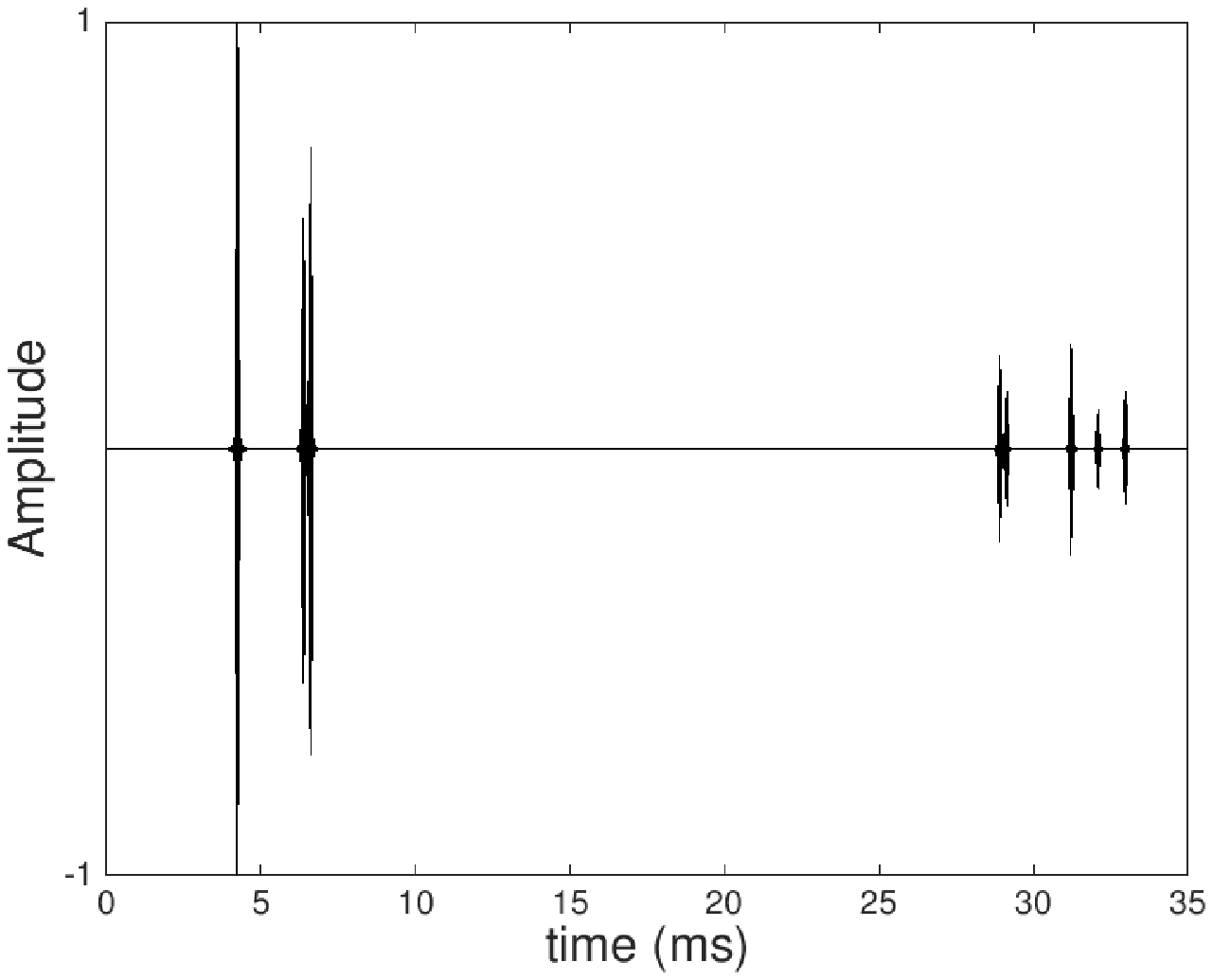}}
        \caption{Impulse response of main lobe at 4 different locations along a circular trajectory.}
        \label{fig9}
\end{figure}

Figure \ref{fig5} presents a tree when the Quad-rotor navigates through the tree and observes no leaf in the sensor main lobe, hence no output is  generated. Figure \ref{fig6} demonstrates the situation when the sonar encounters leaves and branches, which results in impulse as shown in Figure \ref{fig7}. Figure \ref{fig8} presents two trees when the Quad-rotor navigates through the tree in a circular path. It encounters leaves and branches at four instances. The impulse responses of the four sample points are shown in Figure \ref{fig9}. In addition, Figure 10 shows two and three trees with multiple sonar sample points.

\begin{figure}[h]
\hspace{-0.5cm}
  \includegraphics[width=1.2\linewidth]{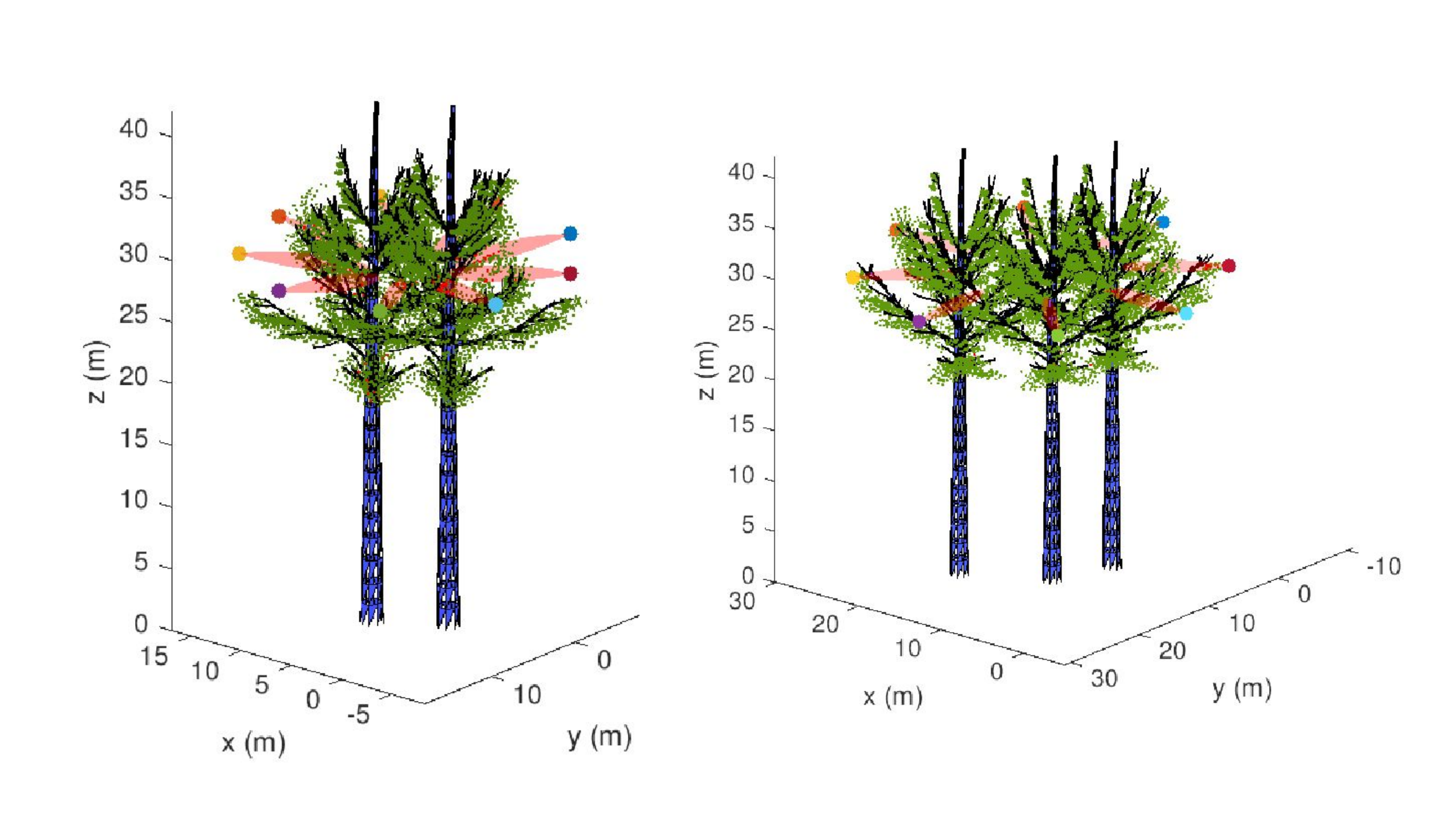}
  \caption{Quad-rotor navigating across different number of trees with leaves encountering in main lobe of sensor}
  \label{fig10}
\end{figure}

%Although Quad-rotor follows a circular path around n number of trees, it can be assumed that several points generate impulses as it encounters branches or leaves in its main lobes as shown in Figure \ref{fig10}. \cmt{(Fix language for this paragraph. Try to use short sentences. One meaning per sentence; and connect the logic.)}

To analyze the computational complexity, we compute the total computation time while the Quad-rotor completes an entire trajectory. Under one setup, we increase the number of locations along the trajectory where the impulse responses are computed. Under another setup, we increase the number of trees (\texttt{T}) in the environment. The computation time for different scenarios are shown in Table~\ref{table1}. We observe that, increasing the number of computation points has a direct effect on the computation time, which is quite intuitive. It is interesting to note that for a circular trajectory around one tree (\texttt{T} = 1) with radius of $6.2$, it takes only around one second on average to compute 15 impulse responses at regular intervals of 24 degrees. For more than one tree, we set the centre of the circular trajectory to be the mean position of the tree locations. We observe that, increasing the number of trees varies the leaf densities and hence has a direct effect on the computation time. For \texttt{T} = 5, it takes only about 3 seconds on average to compute impulses at 15 points along the trajectory. Overall, our model performs fairly well in real-time.

%Hassan's text for table
%It can be seen from results that while Quad-rotor is following a circular path, many sample points can be taken depending upon the computational time as it requires higher mathematical complex solving, discussed in Methodology section. So sample points of main lobe is taken from 1 till 15 while robot is following its trajectory. The different number of trees (\texttt{T}) are considered using IPP equations and the sample points of circular path has been considered, the results of computational time is shown in \ref{table1}. 

\setlength{\tabcolsep}{2.5pt}
	\begin{table}[h]
		%\begin{tabular}{l l @{\hskip .1in}| ccc || ccc}   
		%\scalebox{0.6}
		{ \begin{tabular}{c l  ccccc  }                             
			\hline 	
				\multicolumn{1}{c}{\multirow{1}{*}{Computation points}}    & \multicolumn{5}{c}{Impulse computation time (s)}  \\
								{} & \multicolumn{1}{c}{$\texttt{T}=1$} &  \multicolumn{1}{c}{$\texttt{T}=2$} & \multicolumn{1}{c}{$\texttt{T}=3$} & \multicolumn{1}{c}{$\texttt{T}=4$} &  \multicolumn{1}{c}{$\texttt{T}=5$} &  \\
				
				\hline        
	 1  & 0.73  & 0.77  & 0.86  & 0.93 & 1.07 \\
     5  & 0.88  & 1.00  & 1.19   & 1.33 & 1.52  \\
    10  & 0.97  & 1.21  & 1.42  & 1.72 & 2.11\\
    15  & 1.03  & 1.68  & 1.92   & 2.54 & 3.03 \\
                     \hline		
  		\end{tabular}}                                          
		\caption{Total impulse computation for an entire circular trajectory when the impulses are computed at $1,5,10$ and $15$ locations along the trajectory. \texttt{T} = \{1,\ldots,5\} represent different number of trees in the environment. The center of the circular trajectory is the mean location of the trees in the environment.} 
\label{table1}
	\end{table}

\section{Conclusion}

In this research we explore how to recreate bat behaviors on Quad-rotor UAVs in dynamic environments, thereby transforming nature into bio-technology. In particular, we propose a computational approach to simulate the sensing environments and to simulate foliage echoes during different sensing scenarios.
%to locate forest trees using the Sonar main lob inspired by BATS, installed on the robot Quad-rotor to avoid deviations from their paths and to circumvent obstacles to carry out safe and efficient navigation tasks. Additional impulses are generated between the position of trees and BATS that are encountered in the main lobe of Sonar Beam and in future will be adopted to determine the descriptions of trees. The scenario becomes more intricate for impulse computation as multiple trees are encountered while the Quad-rotor moves dynamically when monitoring the environment. This require a more complex computation to be considered and the results for the has been discussed in this article. Eventually, Quad-rotor determination is processed and tests with several heavily forested trees and Quad-rotors are shown, using navigation targeting information exclusively.

In this preliminary study, we mainly focused on model development and experimental validation in a simulated/known environmental setting. The impulse responses can be further analyzed using state-of-the-art artificial intelligence and machine learning methods to predict different parameters like leaf density, orientation, density. This is a promising direction since it enables navigating in unknown environments. Currently, the trajectories followed by the sonar are predefined and we only analyze the impulse generated at different time instances along the trajectory. Immediate next step is to extend it to an active navigation scenario in which an optimal path can be calculated. %minimizing a certain metric that is a function of the impulses returned. 
Another interesting future direction is to extend the framework towards task and motion planning in large knowledge-intensive domains, as recently done in~\cite{lo2018AAMAS,thomas2019ISRR}. We also plan to model the shading effect between leaves, for example by using an adjusted attenuation function. In order to deal with real world uncertainties, we also plan to integrate our model with Inverse perspective mapping (IPM) approach. This can be done by mounting a camera on the UAV in order to obtain a bird's eye view \cite{10.1007/978-3-030-31993-9_21}.

%\section{ Acknowledgments}
%This research is supported in part by National Science Foundation (NSF) grant #1762577. 

\bibliography{references}
\bibliographystyle{unsrtnat}

\end{document}